\documentclass[letterpaper, 10 pt, conference]{ieeeconf}  

\IEEEoverridecommandlockouts                              

\usepackage{graphics} 
\usepackage{times} 
\usepackage{amsmath} 

\usepackage{amssymb}
\usepackage{makecell}
\usepackage[export]{adjustbox} 
\usepackage{multirow}
\usepackage{subfig}
\usepackage{overpic}
\graphicspath{{images/}}
\usepackage[dvipsnames]{xcolor}
\usepackage[pagebackref,breaklinks,colorlinks]{hyperref}

\definecolor{MyBlack}{HTML}{323A45}

\def\vs{\emph{vs.\ }}

\def\ie{\emph{i.e.\ }}

\def\etc{\emph{etc.\ }}

\def\etal{\emph{et al.\ }}
\def\alambic{\includegraphics[width=0.015\textwidth]{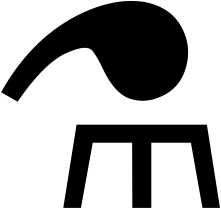}}

\title{\LARGE \bf
LocalViT: Analyzing Locality in Vision Transformers
}

\author{Yawei Li$^{1}$, Kai Zhang$^{1}$, Jiezhang Cao$^{1}$, Radu Timofte$^{1,2}$, Michele Magno$^{3}$, Luca Benini$^{4,5}$, Luc Van Gool$^{1,6}$
\thanks{$^{1}$Computer Vision Lab, D-ITET, ETH Z\"urich, Switzerland. E-mail: {\tt yawei.li@vision.ee.ethz.ch}.}
\thanks{$^{2}$Center for Artificial Intelligence and Data Science (CAIDAS), the University of  W\"urzburg, Germany.}
\thanks{$^{3}$Center for Project-Based Learning, D-ITET, ETH Z\"urich, Switzerland.}
\thanks{$^{4}$Integrated Systems Laboratory, D-ITET, ETH Z\"urich, Switzerland.}
\thanks{$^{5}$Department of Electrical, Electronic and Information Engineering, University of Bologna, Italy.}
\thanks{$^{6}$Processing Speech and Images (PSI), KU Leuven, Belgium.}
}

\begin{document}

\maketitle
\thispagestyle{empty}
\pagestyle{empty}

\begin{abstract}

The aim of this paper is to study the influence of locality mechanisms in vision transformers. Transformers originated from machine translation and are particularly good at modelling long-range dependencies within a long sequence. Although the global interaction between the token embeddings could be well modelled by the self-attention mechanism of transformers, what is lacking is a locality mechanism for information exchange within a local region. 
%
In this paper, locality mechanism is systematically investigated by carefully designed controlled experiments.
We add locality to vision transformers into the feed-forward network. This seemingly simple solution is inspired by the comparison between feed-forward networks and inverted residual blocks.
The importance of locality mechanisms is validated in two ways: 1) A wide range of design choices (activation function, layer placement, expansion ratio) are available for incorporating locality mechanisms and proper choices can lead to a performance gain over the baseline, and 2) The same locality mechanism is successfully applied to vision transformers with different architecture designs, which shows the generalization of the locality concept. For ImageNet2012 classification, the locality-enhanced transformers outperform the baselines Swin-T~\cite{liu2021swin}, DeiT-T~\cite{touvron2020training} and PVT-T~\cite{wang2021pyramid} by 1.0\%, 2.6\% and 3.1\% with a negligible increase in the number of parameters and computational effort. 
Code is available at \url{https://github.com/ofsoundof/LocalViT}.

\end{abstract}

\section{Introduction}
\label{sec:introduction}

Recent advances in machine learning (ML) research, such as computer vision and natural language processing, have been driven by backbone models that can be adapted to different problems~\cite{vaswani2017attention,dosovitskiy2020image}. However, the foundation models in vision and language tend to be heavy and computational expensive, thus hindering their applicability in edge devices such as drones and robotics. Thus, how to improve the efficiency of neural models such as convolutional neural networks (CNNs) and transformers becomes important.


CNNs are based on locality in that convolutional filters only perceive a local region of the input image, \ie the receptive field. By  stacking multiple layers, the effective receptive fields of a deep neural network can be enlarged progressively. This design enables the network to learn a hierarchy of deep features, which is essential for the success of CNNs. Meanwhile, the local, repetitive connections save many parameters compared with fully connected layers.
Yet, one problem is that a larger receptive field can only be achieved by combining layers, despite alternative attempts at enlarging the receptive field~\cite{yu2015multi}.

\begin{figure}[!htb]
    \centering
    \includegraphics[width=1.0\linewidth]{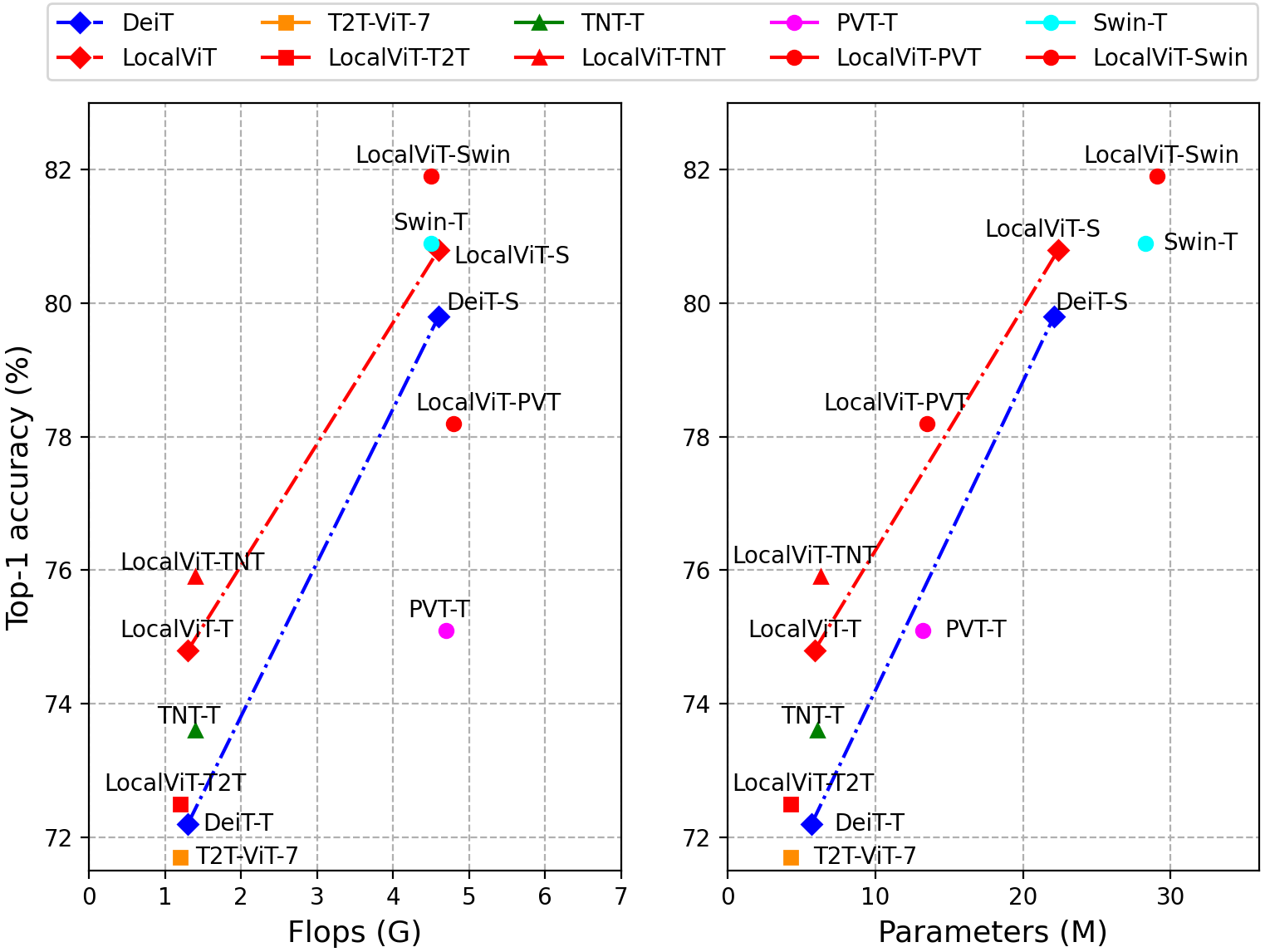}
    \caption{Comparison between LocalViT and the baseline transformers. The transformers enhanced by the proposed locality mechanism outperform their baselines}
    \label{fig:teaser}
\end{figure}

A parallel research strand incorporates global connectivity into the network via self-attention~\cite{vaswani2017attention,wang2020hat,mehta2020delight,wu2020lite,li2023efficient}. This family of networks, \ie transformer networks, originates from machine translation and is very good at modelling long-range dependencies in sequences. There also is a rising interest in applying transformers to vision~\cite{carion2020end,dosovitskiy2020image,touvron2020training}. Vision transformers have already achieved performances quite competitive with their CNN counterparts. 

To process 2D images with transformers, the input image is first converted to a sequence of tokens which correspond to patches in the image. Then the attention module attends to all tokens and a weighted sum is computed as the tokens for the next layer.
In this way, the effective receptive field is expanded to the whole image via a single self-attention layer. Yet, the problem of visual transformers is that global connectivity contradicts the convolutional idea.

Considering the merits of CNNs vs. transformers, a natural question is \textit{whether we can efficiently combine the locality of CNNs and the global connectivity of vision transformers to improve performance while not increasing model complexity.} 

This aim of this paper is aligned with other works that try to answer this interesting question, \ie taking the advantage of both convolution and transformers~\cite{guo2021cmt,liu2021transformer,yuan2021incorporating,srinivas2021bottleneck,li2021bossnas,d2021convit}. Differently, we provide a systematic analysis of various design choices of the locality mechanism by rigorous and controlled experiments. Beyond that, we try to generalize the conclusions from the ablation study to a bunch of vision transformers. Thus, the aim of this paper is to thoroughly investigate a single component (locality mechanism) in vision transformers. 

To conduct the investigation, we start with a mechanism that injects locality into the feed-forward network of transformers, which is inspired by examining the feed-forward network and inverted residuals~\cite{sandler2018mobilenetv2,howard2019searching}. The feed-forward network of transformers consists of two fully connected layers and the hidden dimension between them is expanded to extract richer features. 
Similarly, in inverted residual blocks, the hidden channel between the two $1 \times 1$ convolutions is also expanded. The major difference between them is the efficient depth-wise convolution in the inverted residual block. 
Such convolution can provide precisely the mechanism for local information aggregation which is missing in the feed-forward network of vision transformers.
%
%
To cope with the convolution, the image tokens of the sequence from the self-attention module must be rearranged to a 2D feature map, which is processed by the feed-forward network. The class token is split out and bypasses the feed-forward network. The derived new feature map is converted back to image tokens and concatenated with the bypassed class token. The concatenated sequence is processed by the next transformer layer. 

Through the empirical study, we derive two sets of conclusions.
\textbf{Firstly}, four important properties of the investigated locality mechanism are revealed. 
\textit{i.} Locality mechanism alone can already improve the performance of the baseline transformer. \textit{ii.} A better activation function can result in a significant performance gain. \textit{iii.} The locality mechanism is more important for lower layers. \textit{iv.} Expanding the hidden dimension of the feed-forward network leads to a larger model capacity and a higher classification accuracy. \textbf{Secondly}, as shown in Fig.~\ref{fig:teaser}, the locality mechanism is successfully applied to 5 vision transformers, which underlines its generality.  
The contributions of this paper are three-fold:
\begin{enumerate}
    \item We study a locality mechanism that enhances vision transformers. The modified transformer architecture combines a self-attention mechanism for global relationship modelling and a locality mechanism for local information aggregation.
    \item We analyze the basic properties of the introduced locality mechanism. The influence of each component (depth-wise convolution, non-linear activation function, layer placement, and hidden dimension expansion ratio) is singled out.
    \item We apply these ideas to vision transformers incl. DeiT~\cite{touvron2020training}, Swin transformers~\cite{liu2021swin}, T2T-ViT~\cite{yuan2021tokens}, PVT~\cite{wang2021pyramid}, and TNT~\cite{han2021transformer}. Experiments show that the simple technique proposed in this paper generalizes well to various transformer architectures.
\end{enumerate}

\section{Related Work}
\label{sec:related_works}

\subsection{Transformers}
\label{subsec:transformers}

Transformers were first introduced in~\cite{vaswani2017attention} for machine translation. The proposed attention mechanism aggregates information from the whole input sequence. Thus, transformers are especially good at modelling long-range dependencies between elements of a sequence. Since then, there have been several attempts to adapt transformers towards vision and robotics~\cite{carion2020end,saleh2022cloudattention,brohan2022rt,shi2022recognition,lee2022fully,mascaro2022robust,bucker2022reshaping}. Most recently, transformers are proposed to solve robotic problems by learning the mapping from language and vision observations to robot actions~\cite{brohan2022rt}.

\subsection{Locality \vs global connectivity}

Both local information and global connectivity help to reason about the relationships between  image contents. 
%
The convolution operation applies a sliding window to the input and local information is inherently aggregated to compute new representations. Thus, locality is an intrinsic property of CNNs~\cite{lecun1995convolutional}. 
Although CNNs can extract information from a larger receptive field by stacking layers and forming deep networks, they still lack global connectivity~\cite{krizhevsky2012imagenet,simonyan2014very,he2016deep}. To overcome this problem, some researchers add global connectivity to CNNs with non-local blocks~\cite{wang2018non,liu2018non}.

By contrast, transformers are especially good at modelling long-range dependencies within a a sequence owing to their attention mechanism~\cite{vaswani2017attention}. But, in return, a locality mechanism remains to be added for visual perception. Some works already contributed towards this goal~\cite{yuan2021tokens,liu2021swin,li2021local,wang2021crossformer,chu2021twins,zhang2021multi}. Those works mainly focus on improving the tokenization and self-attention parts. There are also some works that introduce hybrid architectures of CNNs and transformers~\cite{srinivas2021bottleneck,li2021bossnas}.
In the meanwhile, we also noticed some other works introducing convolutions to different parts of transformers~\cite{d2021convit,guo2021cmt}. 
The difference between our work and the other works is that we systematically investigate locality mechanism and single out its importance to transformer architectures. This is inspired by the comparison between vision transformers and the inverted residual blocks in MobileNets. 

\subsection{Inverted residuals}

Compared with normal convolution, the computations of depth-wise convolution are only conducted channel-wise. That is, to obtain a channel of the output feature map, the convolution is only conducted on one input feature map. Thus, depth-wise convolution is efficient both in terms of parameters and computation. Thus, Howard \etal first proposed the MobileNet architecture based on depth-wise separable convolutions~\cite{howard2017mobilenets}. This lightweight and computationally efficient network is quite friendly for mobile devices. Since then, depth-wise convolution has been widely used to design efficient models.
Inverted residual blocks are based on depth-wise convolution and were first introduced in MobileNetV2~\cite{sandler2018mobilenetv2}. 
The inverted residual blocks are composed of a sequence of $1\times1$ - depth-wise -$1\times1$ convolutions. The hidden dimension between the two $1\times1$ convolutions is expanded. The utilization of depth-wise convolution avoids the drastic increase of model complexity brought by normal convolution. Due to the efficiency of this module, it is widely used to form the search space of neural architecture search (NAS)~\cite{howard2019searching,tan2019mnasnet,tan2019efficientnet}. The expansion of the hidden dimension of inverted residuals is quite similar to the feed-forward network of vision transformers. This motivates us to think about the connection between them (See Sec.~\ref{subsec:locality}).

\section{Methodology}
\label{sec:methodology}

Transformers are usually composed of encoders and decoders with similar building blocks. For the image classification task considered here, only the encoders are included in the network. Thus, we mainly describe the operations in the encoder layers. The encoders have two components, \ie the self-attention mechanism that relates a token to all of the tokens and a feed-forward network that is applied to every token. We specifically explain how to introduce locality into the feed-forward network.

\subsection{Input interpretation}
\label{subsec:input_interpretation}

\begin{figure}[!htb]
\centering
\subfloat[][\scriptsize{The input is regarded as a sequence of tokens.
    }]{
    \includegraphics[width=0.95\linewidth]{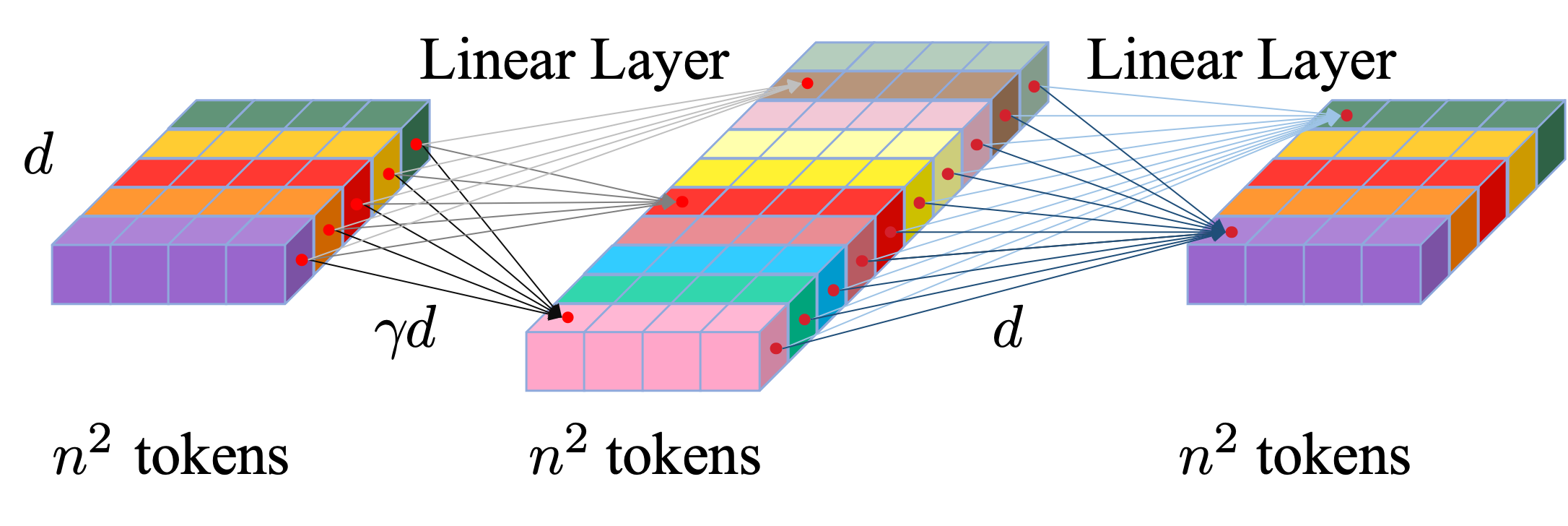}
    \label{fig:feed_forward_sequence}
}

\subfloat[][\scriptsize{An equivalent perspective is to still rearrange the tokens as a 2D lattice.}]{\includegraphics[width=0.95\linewidth]{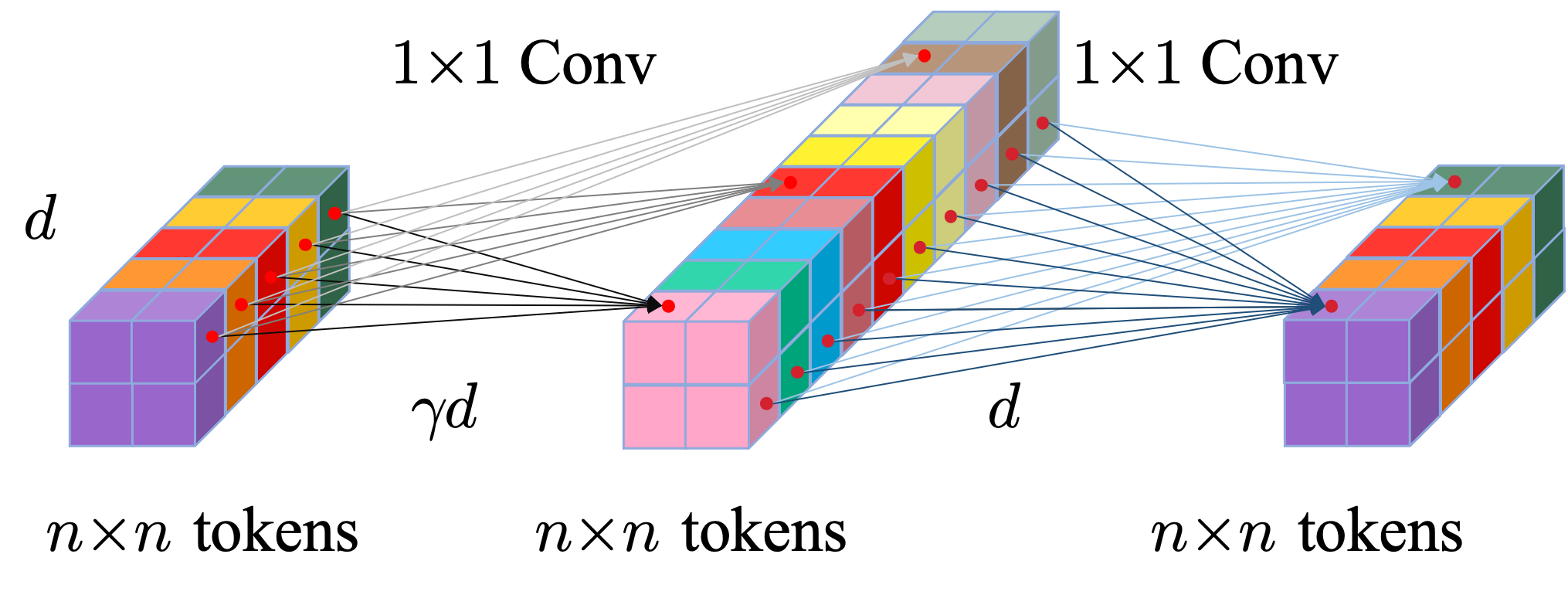}
    \label{fig:feed_forward_lattice}
}
\caption{Visualization of the feed-forward network in transformers from different perspectives. In this figure, $n=2$, $\gamma=2$, $d=5$.}
\label{fig:feed_forward}
\end{figure}

\textbf{Sequence perspective.}
Inherited from language modelling, transformers regard the input as a sequence that contains elements of embedded vectors. Consider an input image $\mathbf{X} \in \mathbb{R}^{C \times H \times W}$, where $C$ and $H \times W$ denote the channel and spatial dimension of the input image, resp. The input image is first converted to tokens $\{\hat{\mathbf{X}}_i \in \mathbb{R}^{d}|i = 1, \ldots, N\}$, where $d = C \times p^2$ is the embedding dimension and $N = \frac{HW}{p^2}$. The tokens can be aggregated into a matrix $\hat{\mathbf{X}} \in \mathbb{R}^{N \times d}$.

\textit{Self-attention.} 
In the self-attention mechanism, the relationship between the tokens is modelled by the similarity between the projected query-key pairs, yielding the attention score. The new tokens are computed as the weighted sum of the project values. That is,
\begin{equation}
    \mathbf{Z} = \mathrm{Softmax}(\mathbf{Q}\mathbf{K}^T / \sqrt{d})\mathbf{V},
    \label{eqn:attention}
\end{equation}
where the $\mathrm{Softmax}$ function is applied to the rows of the similarity matrix and $d$ provides a normalization. 
The query, key, and value are a projection of the tokens, \ie $\mathbf{Q} = \hat{\mathbf{X}}\mathbf{W}_Q$, $\mathbf{K} = \hat{\mathbf{X}}\mathbf{W}_K$, $\mathbf{V} = \hat{\mathbf{X}}\mathbf{W}_V$. The projection matrices $\mathbf{W}_Q$ and $\mathbf{W}_K$ have the same size while $\mathbf{W}_V$ could have a different size. In practice, the three projection matrices usually have the same size, \ie $\mathbf{W}_Q, \mathbf{W}_K, \mathbf{W}_V \in \mathbb{R}^{d\times d}$.

\textit{Feed-forward network.}
After the self-attention layer, a feed-forward network is appended. The feed-forward network consists of two fully-connected layers and transforms the features along the embedding dimension. The hidden dimension between the two fully-connected layers is expanded to learn a richer feature representation. That is,
\begin{equation}
    \mathbf{Y} = f(\mathbf{Z}\mathbf{W}_1)\mathbf{W}_2,
    \label{eqn:feed_forward}
\end{equation}
where $\mathbf{W}_1 \in \mathbb{R}^{d \times \gamma d}$, $\mathbf{W}_2 \in \mathbb{R}^{\gamma d \times d}$, and $f(\cdot)$ denotes a non-linear activation function. For the sake of simplicity, the bias term is omitted. The dimension expansion ratio $\gamma$ is usually set to 4. As shown in Fig.~\ref{fig:feed_forward_sequence}, the input to the feed-forward network is regarded as a sequence of embedding vectors. 

\textbf{Lattice perspective.} Since the feed-forward network is applied position-wise to a sequence of tokens $\mathbf{Z} \in \mathbb{R}^{N \times d}$, an exactly equivalent representation is to rearrange the sequence of tokens into a 2D lattice as shown in Fig.~\ref{fig:feed_forward_lattice}. Then the reshaped feature representation is 
\begin{equation}
    \mathbf{Z}^r = \mathrm{Seq2Img}(\mathbf{Z}), \mathbf{Z}^r \in \mathbb{R}^{h \times w \times d},
    \label{eqn:token_reshape}
\end{equation}
where $h = H/p$ and $w = W / p$.
The operation $\mathrm{Seq2Img}$ converts a sequence to a 2D feature map.
Each token is placed to a pixel location of the feature map.
The benefit of this perspective is that the proximity between tokens is recovered, which provides the chance to introduce locality into the network. The fully-connected layers could be replaced by $1\times1$ convolutions, \ie
\begin{align}
    \mathbf{Y}^r & = f(\mathbf{Z}^r\circledast \mathbf{W}_1^r)\circledast \mathbf{W}_2^r,
    \label{eqn:feed_forward_conv} \\
    \mathbf{Y} & = \mathrm{Img2Seq}(\mathbf{Y}^r),
    \label{eqn:token_reshape_back}
\end{align}
where $\mathbf{W}_1^r \in \mathbb{R}^{d \times \gamma d \times 1 \times1}$ and $\mathbf{W}_2^r \in \mathbb{R}^{\gamma d \times d \times 1 \times1}$ are reshaped from $\mathbf{W}_1$ and $\mathbf{W}_2$ and represent the convolutional kernels. The operation $\mathrm{Img2Seq}$ converts the image feature map back to a token sequence which is used by the next self-attention layer.

\begin{figure}[!htb]
     \begin{center}
    \begin{overpic}[width=0.98\linewidth]{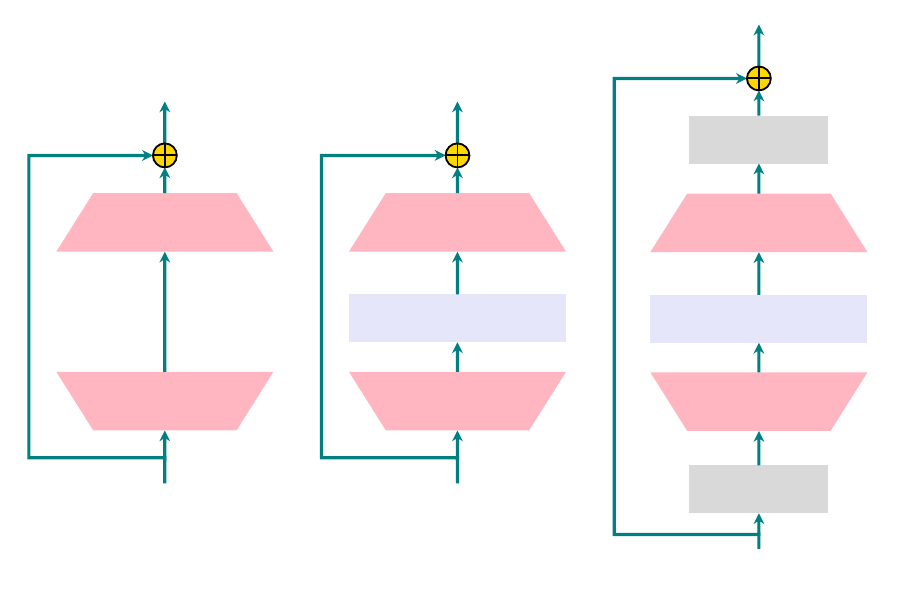}
    \put(10,20){\color{black}{\scriptsize $1\times1$ Conv}}
    \put(10,40){\color{black}{\scriptsize $1\times1$ Conv}}
    
    \put(43,20){\color{black}{\scriptsize $1\times1$ Conv}}
    \put(40,29.2){\color{black}{\scriptsize $3\times3$ DW Conv}}
    \put(43,40){\color{black}{\scriptsize $1\times1$ Conv}}
    
    \put(78.5,10.2){\color{black}{\scriptsize Seq2Img}}
    \put(76.5,20){\color{black}{\scriptsize $1\times1$ Conv}}
    \put(73,29.2){\color{black}{\scriptsize $3\times3$ DW Conv}}
    \put(76.5,40){\color{black}{\scriptsize $1\times1$ Conv}}
    \put(78.5,49.2){\color{black}{\scriptsize Img2Seq}}
    
    \put(16.4,0.3){\color{black}{\scriptsize (a)}}
    \put(49.0,0.3){\color{black}{\scriptsize (b)}}
    \put(82.8,0.3){\color{black}{\scriptsize (c)}}
    
    \end{overpic}
    \end{center}
    \caption{Comparison between the (a) convolutional version of the feed-forward network in vision transformers, the (b) inverted residual blocks, and (c) the utilized network that brings locality mechanism into transformers. ``DW'' denotes depth-wise convolution. To cope with the convolution operation, the conversion between sequence and image feature map is added by ``Seq2Img'' and ``Img2Seq'' in (c). Note that after each convolution, there are activation functions. In Table~\ref{tbl:ablation_activation}, we systematically investigate influence of the activation function after the depthwise convolution in (c).}
    \label{fig:locality}

\end{figure}

\subsection{Locality}
\label{subsec:locality}

Since only $1\times1$ convolution is applied to the feature map, there is a lack of information interaction between adjacent pixels. Besides, the self-attention part of the transformer only captures global dependencies between all of the tokens. Thus, the transformer block does not have a mechanism to model the local dependencies between nearby pixels. It would be interesting if locality could be brought to transformers in an efficient way. 

The expansion of the hidden dimension between fully-connected layers and the lattice perspective of the feed-forward network remind us of the inverted residual block proposed in MobileNets~\cite{sandler2018mobilenetv2,howard2019searching}. As shown in Fig.~\ref{fig:locality}, both of the feed-forward network and the inverted residual expand and squeeze the hidden dimension by $1\times1$ convolution. The only difference is that there is a depth-wise convolution in the inverted residual block. Depth-wise convolution applies a $k \times k$ ($k > 1$) convolution kernel per channel. The features inside the $k \times k$ kernel is aggregated to compute a new feature. Thus, depth-wise convolution is an efficient way of introducing locality into the network. Considering that, we reintroduce depth-wise convolution into the feed-forward network of transformers. And the computation could be represented as   
\begin{equation}
    \mathbf{Y}^r = f\big(f(\mathbf{Z}^r\circledast \mathbf{W}_1^r)\circledast \mathbf{W}_d\big) \circledast \mathbf{W}_2^r,
    \label{eqn:feed_forward_depth}
\end{equation}
where $\mathbf{W}_d \in \mathbb{R}^{\gamma d \times 1 \times k \times k}$ is the kernel of the depth-wise convolution. The finally used network is shown in Fig.~\ref{fig:locality}c. The input, \ie a sequence of tokens is first reshaped to a feature map rearranged on a 2D lattice. Then two $1\times1$ convolutions along with a depth-wise convolution are applied to the feature map. After that, the feature map is reshaped to a sequence of tokens which are used as by the self-attention of the network transformer layer.

Note that the non-linear activation functions are not visualized in Fig.~\ref{fig:locality}. Yet, they play a quite important role in enhancing the network capacity, especially for efficient networks. In particular, we try ReLU6, h-swish~\cite{howard2019searching}, squeeze-and-excitation (SE) module~\cite{hu2018squeeze}, efficient channel attention (ECA) module~\cite{wang2020efficient}, and their combinations. A thorough analysis of the activation function is discussed in the experiments section.

\subsection{Class token}
\label{subsec:class_token}

To apply vision transformers to image classification, a trainable class token is added and inserted into the token embedding, \ie 
\begin{equation}
    \hat{\mathbf{X}} \leftarrow \mathrm{Concat}(\mathbf{X}_{cls}, \hat{\mathbf{X}}),
\end{equation}
where $\leftarrow$ denotes the assignment operation, $\mathbf{X}_{cls} \in \mathbb{R}^{1 \times d}$ is the class token. The new matrix has the dimension of $(N + 1) \times d$ and $N + 1 = \frac{HW}{p^2} + 1$ tokens. In the self-attention module, the class token exchanges information with all other image tokens and gathers information for the final classification. In the feed-forward network, the same transformation is applied to the class and image tokens. 

When depth-wise convolution is introduced into the feed-forward network, the sequence of tokens needs to be rearranged into an image feature map. Yet, the additional dimension brought by the class token makes the exact rearrangement impossible. To circumvent this problem, we split the $N+1$ tokens in Eqn.~\eqref{eqn:attention} into a class token and image tokens again, \ie
\begin{equation}
    (\mathbf{Z}_{cls}, \mathbf{Z}) \leftarrow \mathrm{Split}(\mathbf{Z}).
\end{equation} 
Then the new image token is passed through the feed-forward network according to Eqns.~(\ref{eqn:token_reshape}), (\ref{eqn:feed_forward_depth}), and (\ref{eqn:token_reshape_back}), leading to $\mathbf{Y}$. The class token is not passed through the feed-forward network. Instead, it is directly concatenated with $\mathbf{Y}$, \ie 
\begin{equation}
    \mathbf{Y} \leftarrow \mathrm{Concat}(\mathbf{Z}_{cls}, \mathbf{Y}).
\end{equation}
The split and concatenation of the class token is done for every layer. Although the class token $\mathbf{Z}_{cls}$ is not passed through the feed-forward network, the performance of the overall network is not adversely affected. This is because the information exchange and aggregation is done only in the self-attention part. A feed-forward network like Eqn.~(\ref{eqn:feed_forward}) only enforces a transformation within each token.

\section{Experimental Results}
\label{sec:experimental_results}


This section gives the experimental results for image classification. We first study how the locality brought by depth-wise convolution can improve the performance of transformers. Then we investigate the influence of several design choices including the non-linear activation function, the placement of the locality, and the hidden dimension expansion ratio $\gamma$. All those experiments are based on DeiT-T~\cite{touvron2020training}. Then, we study the generalization to other vision transformers including T2T-ViT~\cite{yuan2021tokens}, PVT~\cite{wang2021pyramid}, TNT~\cite{han2021transformer}, Swin transformer~\cite{liu2021swin} for image classification. The transformers that are equipped with locality are denoted as LocalViT followed by the suffix that denotes the basic architecture. 

\subsection{Implementation details}
\label{subsec:implementation_details}


We introduce the locality mechanism into five vision transformers including DeiT~\cite{touvron2020training}, Swin transformers~\cite{liu2021swin}, T2T-ViT~\cite{yuan2021tokens}, PVT~\cite{wang2021pyramid}, TNT~\cite{han2021transformer}. Since those transformers have different architectures, slightly different considerations should be made. First of all, a Tokens-to-Token (T2T) module is designed in T2T-ViT~\cite{yuan2021tokens} and is inserted into the head of the network. Basically, the T2T module is also a transformer block with feed-forward networks. Thus,  the same modification is also applied to the T2T module. Secondly, TNT introduced an inner transformer block for the image tokens along with the outer transformer block. Yet, we observed huge increase of GPU memory. Thus, the locality mechanism is only applied to the outer transformer block of TNT~\cite{han2021transformer}. Thirdly, for PVT~\cite{wang2021pyramid}, the class token is only introduced in the final stage of the pyramid. Thus, the split and concatenation of the class token for the feed-forward network is only applied in the final stage. 
Fourthly, there is no class token in Swin transformers~\cite{liu2021swin}. The classification is done based on an averaged pooled feature map. 
Thus, the special treatment of the class token is not needed in our modified Swin transformers.


For fast experiment, we shrink TNT and Swin transformers and get smaller versions of them. TNT-T is derived by reducing the embedding dimension from 384 to 192. Swin-M is derived by reducing the number of transformer blocks in the third stage from 6 to 2.

\textbf{Experimental setup.} The ImageNet2012 dataset~\cite{deng2009imagenet} is used in this paper. The dataset contains 1.28M training images and 50K validation images from one thousand classes. We follow the same training protocol as DeiT~\cite{touvron2020training}.
The input image is randomly cropped with size $224 \times 224$. Cross-entropy is used as the loss function. Label smoothing is used. The weight decay factor is set to 0.05. The AdamW optimizer is used with a momentum of 0.9. The training continues for 300 epochs. The batch size is set to 1024. The initial learning rate is set to $1\times10^{-3}$
and decreases to $1\times10^{-5}$ following a cosine learning rate scheduler. During validation, a center crop of the validation images is conducted.
We use 8 NVIDIA TITAN RTX GPUs to run the experiments.

\subsection{Influence of the locality}

\begin{table}[!htb]
    \centering
    \caption{Investigation of the locality brought by depth-wise convolution. *ReLU6 is used as the activation function after depth-wise convolution. \textdagger Results derived by modifying the DeiT architecture and training the network with the same training protocol}
    \label{tbl:ablation_locality}
    \begin{tabular}{l|c|c|c|l}
        \hline
        \multirow{2}{*}{Network} & \multirow{2}{*}{$\gamma$} & \multirow{2}{*}{\makecell{Depthwise \\ Conv}} & Params  & \multicolumn{1}{c}{Top-1} \\ 
        & & & \multicolumn{1}{c|}{(M)}  & \multicolumn{1}{c}{Acc. (\%)} \\\hline
        DeiT-T~\cite{touvron2020training}   & 4 & No & 5.7  & 72.2 \\
        LocalViT-T  & 4 & No  & 5.7  & 72.5 (\textcolor{OrangeRed}{0.3$\uparrow$}) \\
        LocalViT-T* & 4 & Yes & 5.8  & 73.7 (\textcolor{OrangeRed}{1.5$\uparrow$}) \\ \hline
        DeiT-T~\cite{touvron2020training}   & 6 & No & 7.5 & 73.1\textdagger \\
        LocalViT-T  & 6 & No  & 7.5  & 74.3 (\textcolor{OrangeRed}{1.2$\uparrow$}) \\
        LocalViT-T* & 6 & Yes & 7.7  & 76.1 (\textcolor{OrangeRed}{3.0$\uparrow$}) \\ \hline
    \end{tabular}
\end{table}

We first study how the local information could help to improve the performance of vision transformers in Table~\ref{tbl:ablation_locality}. Different hidden dimension expansion ratios $\gamma$ are investigated. First of all, due to the change of the operations in the feed-forward network (Sec.~\ref{subsec:implementation_details}), the Top-1 accuracy of LocalViT-T is slightly increased even without the depth-wise convolution. The performance gain is 0.3\% for $\gamma = 4$ and is increased to 1.2\% for $\gamma = 6$. Note that compared with DeiT-T, no additional parameters and computation are introduced for the improvement. When locality is incorporated into the feed-forward network, there is a significant improvement of the model accuracy, \ie 1.5\% for $\gamma = 4$ and 3.0\% for $\gamma = 6$. Compared with the baseline, there only is a marginal increase in the number of parameters and a negligible increase in the amount of computation. \textit{Thus, the performance of vision transformers can be significantly improved by the incorporation of a locality mechanism and the adaptation of the operation in the feed-forward network.}

\subsection{Activation functions}

\begin{table}[!tb]
    \centering
    \caption{Investigation of the non-linear activation function. The combination of HS, ECA~\cite{wang2020efficient}, and SE~\cite{hu2018squeeze} is studied. ``HS'' means h-swish activation. ``SE-**'' means the reduction ratio in the squeeze-and-excitation module. $\gamma$ is set to 4.}
    \label{tbl:ablation_activation}
    \begin{tabular}{l|c|l}
        \hline
        \multirow{2}{*}{Activation}  & Params  & \multicolumn{1}{c}{Top-1} \\ 
        & \multicolumn{1}{c|}{(M)} & \multicolumn{1}{c}{Acc. (\%)} \\\hline
        Deit-T~\cite{touvron2020training} & 5.7  & 72.2 \\ \hline
        ReLU6           & 5.8  & 73.7 (\textcolor{OrangeRed}{1.5$\uparrow$})\\
        HS         & 5.8  & 74.4 (\textcolor{OrangeRed}{2.2$\uparrow$})\\
        HS + ECA   & 5.8  & 74.5 (\textcolor{OrangeRed}{2.3$\uparrow$})\\
        HS + SE-192& 5.9  & 74.8 (\textcolor{OrangeRed}{2.6$\uparrow$})\\
        HS + SE-96 & 6.0  & 74.8 (\textcolor{OrangeRed}{2.6$\uparrow$})\\
        HS + SE-48 & 6.1  & 75.0 (\textcolor{OrangeRed}{2.8$\uparrow$})\\
        HS + SE-4  & 9.4  & 75.8 (\textcolor{OrangeRed}{3.6$\uparrow$})\\ \hline
         
    \end{tabular}
\end{table}

The non-linear activation function after depth-wise convolution used in the above experiments is simply ReLU6. The benefit of using other non-linear activation functions is also studied. In Table~\ref{tbl:ablation_activation}, the ablation study based on LocalViT-T is done. First of all, by replacing the activation function from ReLU6 to h-swish, the gain of Top-1 accuracy over the baseline is increased from 1.5\% to 2.2\%. This shows the benefit of h-swish activation functions can be easily extended from CNNs to vision transformers. Next, the h-swish activation function is combined with other channel attention modules including ECA~\cite{wang2020efficient} and SE~\cite{hu2018squeeze}. The ECA and SE modules are placed directly after the h-swish function. By adding ECA, the performance is further improved by 0.1\%. Considering that only 60 parameters are introduced, this improvement is still considerable under a harsh parameter budget.

Another significant improvement is brought by a squeeze-and-excitation module. When the reduction ratio in the squeeze-and-excitation module is reduced from 192 to 4, the gain of Top-1 accuracy is gradually increased from 2.6\% to 3.6\%. The number of parameters is also increased accordingly. Note that, for all of the networks, the computational complexity is almost the same. This implies that if there is no strict limitation on the number of parameters, advanced non-linear activation functions could be used. In the following experiments, we use the combination of h-swish and SE as the non-linear activation function after depth-wise convolution. Additionally, the reduction ratio of the squeeze-and-excitation module is chosen such that only 4 channels are kept after the squeeze operation. This choice of design achieves a good balance between the number of parameters and the model accuracy.
\textit{Thus, local information is also important in vision transformers. A wide range of efficient modules could be introduced into the feed-forward network of vision transformers to expand the network capacity.}

\subsection{Placement of locality, expansion ratio, and discussion}
\label{subsec:placement_expansion_discussion}

\begin{table}[!tb]
    \centering
    \caption{Influence of the placement of locality. ``All'' means all of the transformer layers are enhanced by depth-wise convolution. ``Low'', ``Mid'', and ``High'' mean the lower, middle, and higher transformer layers are equipped with depth-wise convolution, respectively. The study is based on LocalViT-T.}
    \label{tbl:ablation_placement}
    \begin{tabular}{c|c|c|c}
        \hline
         Layer & Params & FLOPs & \multicolumn{1}{c}{Top-1} \\ 
         Placement & \multicolumn{1}{c|}{(M)} & \multicolumn{1}{c|}{(G)} & \multicolumn{1}{c}{Acc. (\%)} \\\hline
        High: 9$\sim$12 & 5.78 & 1.26 & 69.1 \\ 
        Mid: 5$\sim$8   & 5.78 & 1.26 & 72.1 \\ 
        Low: 1$\sim$4   & 5.78 & 1.26 & 73.1 \\ 
        Low: 1$\sim$8   & 5.84 & 1.27 & 74.0 \\ 
        All: 1$\sim$12  & 5.91 & 1.28 & 74.8 \\ \hline
    \end{tabular}
\end{table}

\begin{table}[!tb]
    \centering
    \caption{Investigating the expansion ratio of hidden layers in the feed-forward network.}
    \label{tbl:ablation_expansion_ratio}
    \begin{tabular}{c|c|c|c|c}
        \hline
        \multirow{2}{*}{$\gamma$} & \multirow{2}{*}{SE} & Params & FLOPs & \multicolumn{1}{c}{Top-1} \\ 
         & & \multicolumn{1}{c|}{(M)} & \multicolumn{1}{c|}{(G)} & \multicolumn{1}{c}{Acc. (\%)} \\\hline
        \multirow{2}{*}{1} & No & 3.1 & 0.7 & 65.9\\
        & Yes & 3.1 & 0.7 & 66.2\\ \hline
        \multirow{2}{*}{2} & No & 4.0 & 0.9 & 70.1 \\
        & Yes & 4.0 & 0.9 & 70.6 \\ \hline
        \multirow{2}{*}{3} & No & 4.9 & 1.1 & 72.9 \\
        & Yes & 5.0 & 1.1 & 73.1 \\ \hline
        \multirow{2}{*}{4} & No & 5.8 & 1.3 & 74.4 \\
        & Yes & 5.9 & 1.3 & 74.8\\ \hline
    \end{tabular}
\end{table}

The transformer layers where the locality is introduced can also influence the performance of the network. Thus, an ablation study based on LocalViT-T is conducted to study their effect. The results is reported in Table~\ref{tbl:ablation_placement}. There are in total 12 transformer layers in the network. We divide the 12 layers into 3 groups corresponding to ``Low'', ``Mid'', and ``High'' stages. For the former 3 rows of Table~\ref{tbl:ablation_placement}, we independently insert locality into the three stages. As the locality is moved gradually from lower stages to the higher stages, the accuracy of the network is decreased. This shows that local information is especially important for the lower layers. This is also consistent with our intuition. When the depth-wise convolution is applied to the lower layers, the local information aggregated there could also be propagated to the higher layers, which is important to improve the overall performance of the network.

When the locality is introduced only in the higher stage, the Top-1 accuracy is even lower than DeiT-T. To investigate whether locality in the higher layers always has an adverse effect, we progressively allow more lower layers to have depth-wise convolution until locality is enabled for all layers. This corresponds to the last three rows of Table~\ref{tbl:ablation_placement}. Starting from the lower layers, the performance of the network could be gradually improved as locality is enabled for more layers. \textit{Thus, introducing the locality to the lower layers is more advantageous compared with higher layers.}


The effect of the expansion ratio of the hidden dimension of the feed-forward network is also investigated. The results are shown in Table~\ref{tbl:ablation_expansion_ratio}. Expanding the hidden dimension of the feed-forward network can have a significant effect on the performance of the transformers. As the expansion ratio is increased from 1 to 4, the Top-1 accuracy is increased from less than 70\% to nearly 75\%. The model complexity is also almost doubled. 
\textit{Thus, the network performance and model complexity can be balanced by the hidden dimension expansion ratio $\gamma$. Squeeze-and-excitation can be more beneficial for smaller $\gamma$.}

\begin{table}[!t]
    \centering
    \caption{Image classification results for different CNNs and vision transformers. The locality functionality is enabled for five different vision transformers}
    \label{tbl:generalization_results}
    \begin{tabular}{l|r|r|l|c}
        \hline
         Network & \makecell{Params\\(M)} & \makecell{FLOPs\\(G)} & \makecell{Top-1 \\ Acc. (\%)} & \makecell{Top-5 \\ Acc. (\%)} \\
         \hline
         \multicolumn{5}{c}{CNNs} \\ \hline
         ResNet-18~\cite{he2016deep}     & 11.7 & 1.8 & 69.8 & 89.1 \\
         ResNet-50~\cite{he2016deep}     & 25.6 & 4.1 & 76.1 & 92.9 \\
         DenseNet-169~\cite{huang2017densely}         & 14.2 & 3.4 & 75.6 & 92.8\\
         RegNet-4GF~\cite{radosavovic2020designing}   & 20.7 & 4.0 & 80.0 & -- \\
         MobileNetV1~\cite{howard2017mobilenets}      & 4.2 & 0.6 & 70.6 & -- \\
         {MobileNetV2~\cite{sandler2018mobilenetv2}}   & 6.9 & 0.6 & 74.7 & -- \\
         EfficientNet-B0~\cite{tan2019efficientnet}   & 5.3 & 0.4 & 77.1 & 93.3 \\ \hline
         \multicolumn{5}{c}{Transformers} \\ \hline
         DeiT-T~\cite{touvron2020training}     &5.7 & 1.3 &72.2 & 91.1\\
         LocalViT-T     & 5.9 & 1.3 & 74.8 (\textcolor{OrangeRed}{2.6$\uparrow$}) & 92.6 \\
         DeiT-T$\alambic$~\cite{touvron2020training}     & 5.9 & 1.3 & 74.5 & --\\ \hline
         DeiT-S~\cite{touvron2020training}    & 22.1 & 4.6 &79.8 &95.1\\
         LocalViT-S     & 22.4 & 4.6 & 80.8 (\textcolor{OrangeRed}{1.0$\uparrow$}) &95.4\\ 
         DeiT-S$\alambic$~\cite{touvron2020training}     & 22.4 & 4.6 & 81.2 &--\\  \hline
         T2T-ViT-7~\cite{yuan2021tokens}      & 4.3 & 1.2 & 71.7 & --\\
         LocalViT-T2T   & 4.3 & 1.2 & 72.5 (\textcolor{OrangeRed}{0.8$\uparrow$}) &-- \\ \hline
         TNT-T~\cite{han2021transformer}      & 6.1 & 1.4 & 73.6 & 91.9 \\
         LocalViT-TNT   & 6.3 & 1.4 & 75.9 (\textcolor{OrangeRed}{2.3$\uparrow$}) & 93.0 \\ \hline
         PVT-T~\cite{wang2021pyramid}         & 13.2 & 4.7 & 75.1 & 92.3\\ 
         LocalViT-PVT   & 13.5 & 4.8 & 78.2 (\textcolor{OrangeRed}{3.1$\uparrow$}) & 94.2 \\ \hline
         Swin-M~\cite{liu2021swin}         & 21.2 & 3.0 & 79.2 & 94.5 \\ 
         LocalViT-Swin-M   & 21.7 & 3.0 & 80.4 (\textcolor{OrangeRed}{1.2$\uparrow$}) & 95.0 \\
         Swin-T~\cite{liu2021swin}         & 28.3 & 4.5 & 80.9 & 95.3 \\ 
         LocalViT-Swin   & 29.1 & 4.5 & 81.9 (\textcolor{OrangeRed}{1.0$\uparrow$}) & 95.7 \\ \hline
    \end{tabular}
\end{table}

\begin{figure*}[!htb]
\centering
\subfloat[][\scriptsize{DeiT-S~\cite{touvron2020training} \vs LocalViT-S.}]{\includegraphics[trim={17 20 30 20},clip,width=0.24\textwidth,valign=t]{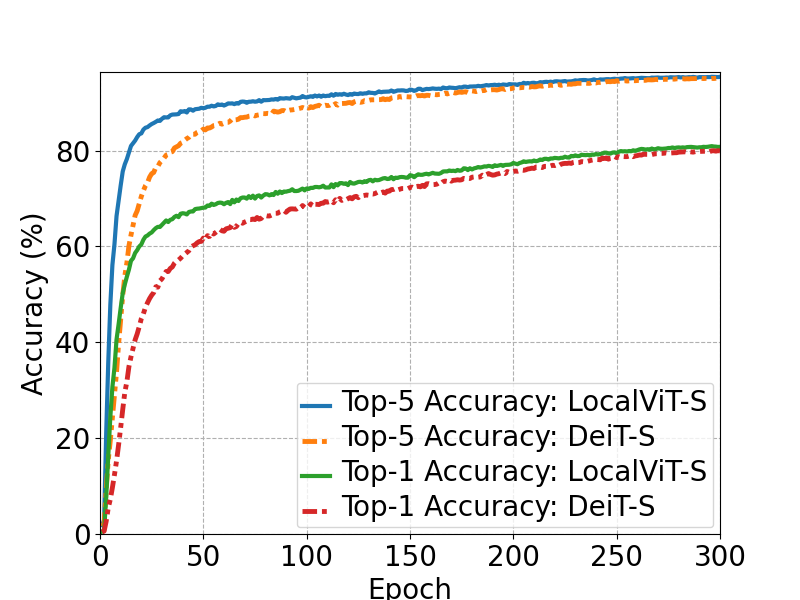}
    \label{fig:accuracy2}
}
%
\subfloat[][\scriptsize{Swin-T~\cite{liu2021swin} \vs LocalViT-Swin.}]{
    \includegraphics[trim={17 20 30 20},clip,width=0.24\textwidth,valign=t]{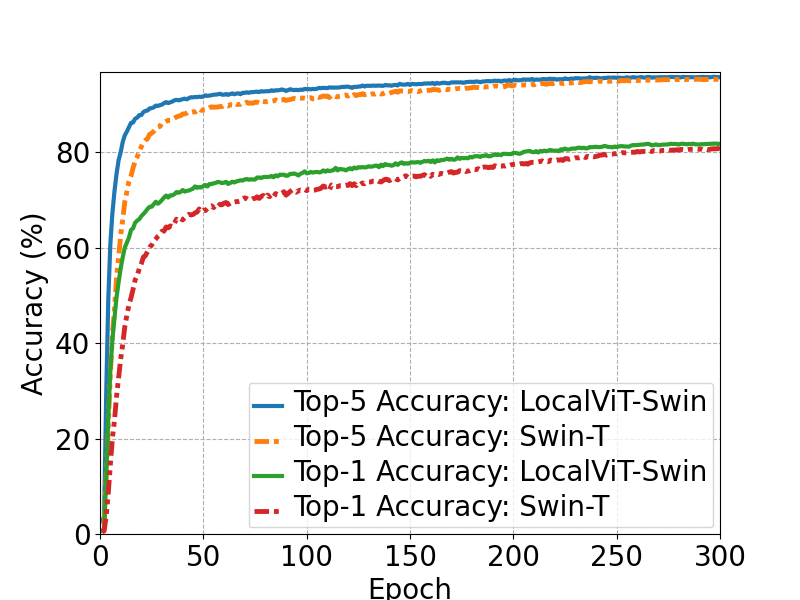}  
    \label{fig:accuracy4}
}
\subfloat[][\scriptsize{PVT-T~\cite{wang2021pyramid} \vs LocalViT-PVT.}]{
    \includegraphics[trim={17 20 30 20},clip,width=0.24\textwidth,valign=t]{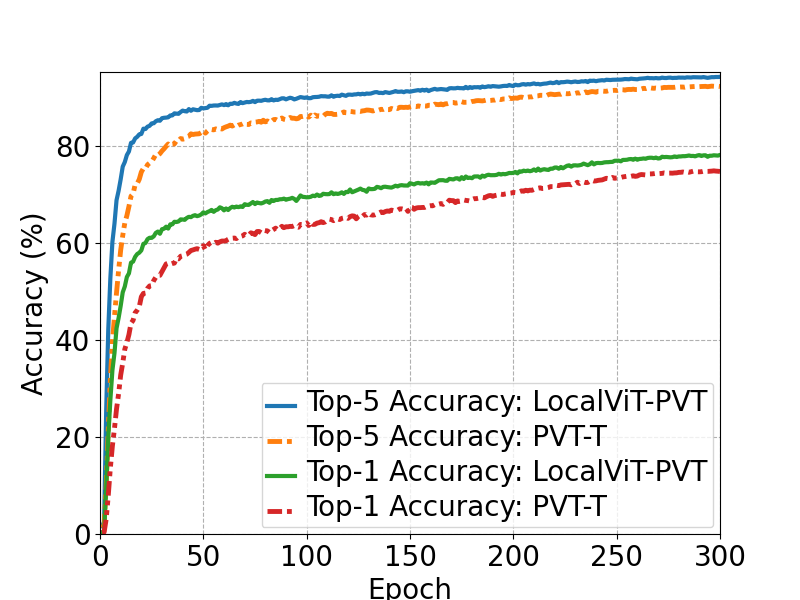} 
    \label{fig:accuracy5}
}
\subfloat[][\scriptsize{TNT-T~\cite{han2021transformer} \vs LocalViT-TNT.}]{\includegraphics[trim={17 20 30 20},clip,width=0.24\textwidth,valign=t]{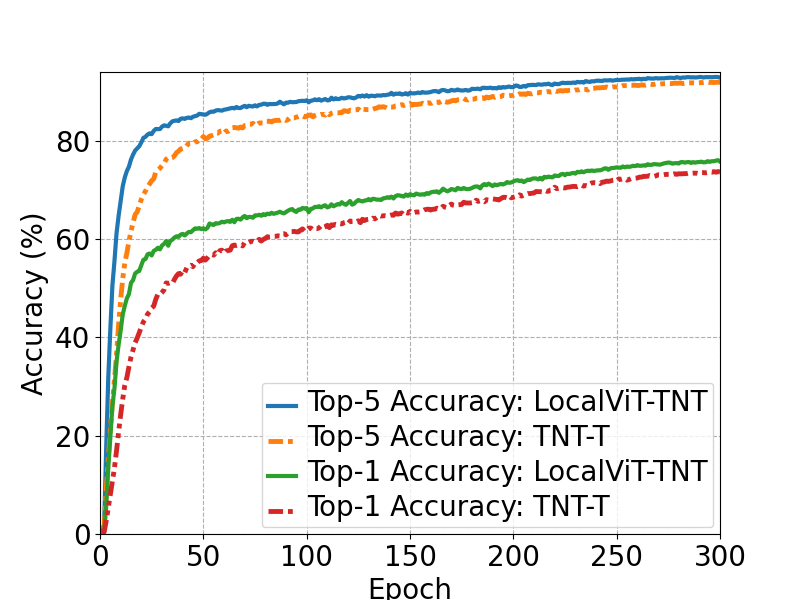}
    \label{fig:accuracy6}
}
\caption{Comparison of Top-1 and Top-5 accuracy between the baseline transformers and the locality enhanced LocalViT}
\label{fig:accuracy}
\end{figure*}



\begin{figure}[!htb]
    \centering
    \subfloat[][\scriptsize Input.]{\includegraphics[width=0.24\linewidth]{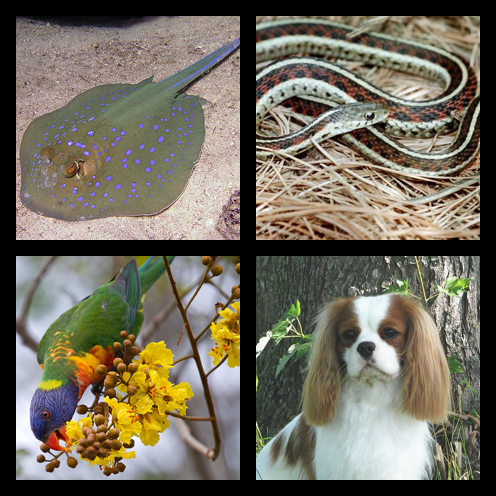}
        \label{fig:input}
    }
    \subfloat[][\scriptsize Pooling.]{\includegraphics[width=0.24\linewidth]{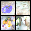}
        \label{fig:pool_input}
    }
    \subfloat[][\scriptsize DeiT]{\includegraphics[width=0.24\linewidth]{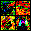}
        \label{fig:feature_map_deit}
    }
    \subfloat[][\scriptsize LocalViT]{\includegraphics[width=0.24\linewidth]{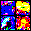}
        \label{fig:feature_map_localvit}
    }
    \caption{Comparison of feature maps. (a) Random sampled input images from ImageNet (b) Max pooled images with kernel size $16 \times 16$. (c) \& (d) Feature map of the last transformer layer from DeiT and LocalViT}
    \label{fig:feature_map}
\end{figure}

\subsection{Generalization and comparison with state-of-the-art}

Finally, we try to incorporate locality into vision transformers including DeiT~\cite{touvron2020training}, Swin transformers~\cite{liu2021swin}, T2T-ViT~\cite{yuan2021tokens}, TNT~\cite{han2021transformer}, PVT~\cite{wang2021pyramid}
and compare their performance with CNNs. 
We draw three major conclusions from Table~\ref{tbl:generalization_results}. \textit{Firstly, the effectiveness of locality can be generalized to a wide range of vision transformers based on the following observations.} \textbf{\textit{1)}} Compared with DeiT, LocalViT can yield a higher classification accuracy for both the tiny and small version of the network. The increase of Top-1 accuracy is 2.6\% and 1.0\%, resp. LocalViT-T even outperforms DeiT-T$\alambic$ which is enhanced by knowledge distillation from RegNetY-160~\cite{radosavovic2020designing}. The small version LocalViT-S is slightly worse than DeiT-S$\alambic$ by 0.4\%. \textbf{\textit{2)}} LocalViT-T2T outperforms T2T-ViT-7 by 0.8\%. Note that T2T-ViT already tries to model the local structure information in the tokens-to-token module. \textbf{\textit{3)}} In TNT, an additional transformer block is used to extract local features for the image tokens. Thus, the locality is also considered in TNT. The modified network, \ie LocalViT-TNT could still improve the classification accuracy by a large margin of 2.3\%. \textbf{\textit{4)}} The biggest improvement comes from PVT. Introducing the locality module leads to a gain of 3.1\% over PVT-T. \textbf{\textit{5)}} Swin transformer already adopts shifted windows that constrain attention in a local region. Yet, adding locality processing module into the network could still improve the performance of Swin transformers.  \textbf{\textit{6)}} The comparison of the training log between the baseline transformers and LocalViT is shown in Fig.~\ref{fig:accuracy}. 
As shown in Fig.~\ref{fig:accuracy}, during the training phase, LocalViT outperforms the baseline transformers consistently in terms of both Top-1 and Top-5 accuracy. The gap between LocalViT and the baseline transformers is more obvious in the early training phase. For example, the gap of the Top-1 accuracy between LocalViT-T and DeiT-T could be as large as 10\% during the early training phase (at about Epoch 25). 
Thus, this confirms that the locality mechanism introduced by LocalViT can enlarge the capacity of vision transformers and lead to better performances.  

\textit{Secondly, some versions of the enhanced vision transformer LocalViT are already quite comparable or even outperform CNNs.} This conclusion can be drawn by making the pairwise comparison, \ie LocalViT-T \vs MobileNetV2 (1.4), LocalViT-S \vs ResNet-50, LocalViT-T2T \vs MobileNetV1, LocalViT-PVT \vs DenseNet-169 \etc

\textit{Thirdly, by comparing the feature maps of transformers with and without locality mechanism in Fig.~\ref{fig:feature_map_deit} and Fig.~\ref{fig:feature_map_localvit}, it is clear that LocalViT does a better job at localizing the objects in the presented input images.}



\textbf{Discussion of limitation.} As shown in Table~\ref{tbl:generalization_results}, introducing locality mechanism increases the complexity of the network. 
As a result, the inference of the network could be slowed down. We report the throughput of different methods on one NVIDIA TITAN Xp GPU in Table~\ref{tbl:throughput}. When comparing with the DeiT-T and TNT-T, there is only a marginal decrease (less than 10\%) of throughput for LocalViT. The throughput of LocalViT-T and DeiT-T is almost comparable. 
For PVT-T and Swin-T, the locality enhanced transformer LocalViT is faced with a larger throughput decrease. But the the decrease is still within 24\%.
Yet, considering the non-trivial improvement of the locality enhanced transformer network, we think this is acceptable. The value of this paper is to show the importance of locality mechanism in vision transformers rather than achieving state-of-the-art performances. 
And we believe that the conclusion derived under rigorous and controlled experiments could help the community to understand the locality mechanism in vision transformers. 
Combination with other design choices towards more efficient networks could be done in follow-up works.


\begin{table}[!htb]
    \centering
    \caption{Throughput comparison between the baseline networks and those enhanced by the locality mechanism. The inference is conducted on a single NVIDIA TITAN Xp GPU}    
    \begin{tabular}{@{}p{0.11\textwidth}@{}p{0.11\textwidth}||p{0.11\textwidth}@{}p{0.11\textwidth}@{}}
            \hline
            Network  & \makecell{Throughput \\ images/s }& Network  & \makecell{Throughput \\ images/s } \\\hline
            DeiT-T          & 346.3 & PVT-T           & 318.4\\ 
            LocalViT-T      & 336.9 ({2.7\% $\downarrow$}) & LocalViT-PVT    & 248.3 ({22.0\% $\downarrow$}) \\ \hline
            TNT-T           & 222.0 & Swin-T          & 206.3\\
            LocalViT-TNT    & 208.3 ({6.2\% $\downarrow$}) & LocalViT-Swin   & 158.2 ({23.3\% $\downarrow$}) \\ \hline
        \end{tabular}
    \label{tbl:throughput}
\end{table}


\section{Conclusion}
\label{sec:conclusion}

In this paper, we investigated the influence of locality mechanism in the feed-forward of vision transformers. We introduced the locality mechanism into vision transformers by incorporating 2D depth-wise convolutions followed by a non-linear activation function into the feed-forward network of vision transformers. 
To cope with the locality mechanism, the sequence of tokens embedding is rearranged into a lattice as a 2D feature map, which is used as the input to the enhanced feed-forward network. To enable the rearrangement, the class token is split before the feed-forward network and concatenated with other image embeddings after the feed-forward network. A series of studies were made to investigate various factors (activation function, layer placement, and expansion ratio) that might influence of performance of the locality mechanism. The proposed locality mechanism is successfully applied to five different vision transformers, which validates its generality.






\bibliographystyle{IEEEtran}
\bibliography{main}



\end{document}